\documentclass{article}

\usepackage[preprint]{neurips_2024}

\usepackage[utf8]{inputenc} % allow utf-8 input
\usepackage[T1]{fontenc}    % use 8-bit T1 fonts
\usepackage{hyperref}       % hyperlinks
\usepackage{url}            % simple URL typesetting
\usepackage{booktabs}       % professional-quality tables
\usepackage{amsfonts}       % blackboard math symbols
\usepackage{nicefrac}       % compact symbols for 1/2, etc.
\usepackage{microtype}      % microtypography
\usepackage{xcolor}         % colors
\usepackage{amsmath,graphicx}
\usepackage{bm}
\usepackage{multirow}
\usepackage{arydshln}

\title{MERaLiON-SpeechEncoder: Towards a Speech Foundation Model for Singapore and Beyond}

\author{\large{MERaLiON Team} \vspace{4mm}\\
    \textbf{Muhammad Huzaifah}\thanks{Core contributors listed by alphabetical order. Please cite this report as authored by MERaLiON Team.
\\Correspondence: \{huzaifah\_md\_shahrin, sailor\_hardik\_bhupendra\}@i2r.a-star.edu.sg}\enspace, \textbf{Geyu Lin}$^{\ast}$, \textbf{Tianchi Liu}$^{\ast}$, \textbf{Hardik B. Sailor}$^{\ast}$, \textbf{Kye Min Tan}$^{\ast}$\\
    \textbf{Tarun K. Vangani}$^{\ast}$, \textbf{Qiongqiong Wang}$^{\ast}$, \textbf{Jeremy H. M. Wong}$^{\ast}$, \textbf{Jinyang Wu}$^{\ast}$\vspace{1mm} \\
    \textbf{Nancy F. Chen}, \textbf{Ai Ti Aw} \vspace{2mm}\\
    Institute for Infocomm Research (I$^2$R), A*STAR, Singapore \\ }

\begin{document}

\maketitle

\begin{abstract}
  % The abstract paragraph should be indented \nicefrac{1}{2}~inch (3~picas) on
  % both the left- and right-hand margins. Use 10~point type, with a vertical
  % spacing (leading) of 11~points.  The word \textbf{Abstract} must be centered,
  % bold, and in point size 12. Two line spaces precede the abstract. The abstract
  % must be limited to one paragraph.
This technical report describes the MERaLiON-SpeechEncoder, a foundation model designed to support a wide range of downstream speech applications. Developed as part of Singapore's National Multimodal Large Language Model Programme, the MERaLiON-SpeechEncoder is tailored to address the speech processing needs in Singapore and the surrounding Southeast Asian region. The model currently supports mainly English, including the variety spoken in Singapore. We are actively expanding our datasets to gradually cover other languages in subsequent releases. The MERaLiON-SpeechEncoder was pre-trained from scratch on 200,000 hours of unlabelled speech data using a self-supervised learning approach based on masked language modelling. We describe our training procedure and hyperparameter tuning experiments in detail below. Our evaluation demonstrates improvements to spontaneous and Singapore speech benchmarks for speech recognition, while remaining competitive to other state-of-the-art speech encoders across ten other speech tasks. We commit to releasing our model, supporting broader research endeavours, both in Singapore and beyond.
\end{abstract}

\setcounter{footnote}{0}% Reset footnote counter

\section{Introduction}
We present the MERaLiON\footnote{Multimodal Empathetic Reasoning and Learning in One Network. This name is a reference to our AudioLLM under development and is co-opted for adjacent models under the National Multimodal Large Language Model Programme \citep{LLM-prog}.} -SpeechEncoder, a 630M parameter speech foundation model pre-trained on large-scale data to support a wide variety of downstream speech applications. Pre-training was carried out from scratch, under a self-supervised learning (SSL) framework, utilising a BERT-like masked language modelling objective. The current model supports primarily English, with a particular focus on Singapore-accented English and the English-based creole \textit{Singlish}, which includes and is influenced by a mix of other languages, including Hokkien, Malay, Cantonese, and Tamil. Our goal is to gradually support other major languages spoken throughout Southeast Asia in subsequent releases.

While the encoder can be used stand-alone, development was carried out concurrently with the MERaLiON-AudioLLM \citep{meralion-audiollm}, a multimodal large language model (LLM) that can handle both speech and text inputs, with the objective of eventually integrating the speech encoder within the AudioLLM framework. The integrated model is under active development and will be included in future releases. We outline our main contributions as follows:
\begin{enumerate}
  \item Independent implementation and training of a speech encoder with the BERT-based speech pre-training with random-projection quantizer (BEST-RQ) objective \citep{bestrq}. Given that previous implementations utilising this objective have been closed-source or scaled-down, we fill this gap by making our model checkpoints available via Hugging Face: \url{https://huggingface.co/MERaLiON/MERaLiON-SpeechEncoder-v1}. The checkpoints and finetuning recipes will be useful for developers who want to use the MERaLiON-SpeechEncoder to build their own speech-based systems or for academics who want to explore SSL models. 
  \item Demonstration of our model's strong performance in Singapore English while retraining good ability in general English speech recognition benchmarks, on par with state-of-the-art (SOTA) models.
  \item Evaluation of speech representations trained with the random projection quantiser on a wide range of downstream speech tasks, including ones based on speaker and paralinguistics, as encapsulated by the widely adopted SUPERB benchmark \citep{yang21c_interspeech}. Evaluation outside of automatic speech recognition (ASR) has previously been limited.
  \item Technical sharing of large-scale speech pre-training on an AMD GPU cluster, the first discussion of such an attempt.
\end{enumerate}

\section{Background}

\begin{figure}[htb]
  \centering
  \includegraphics[width=0.92\linewidth]{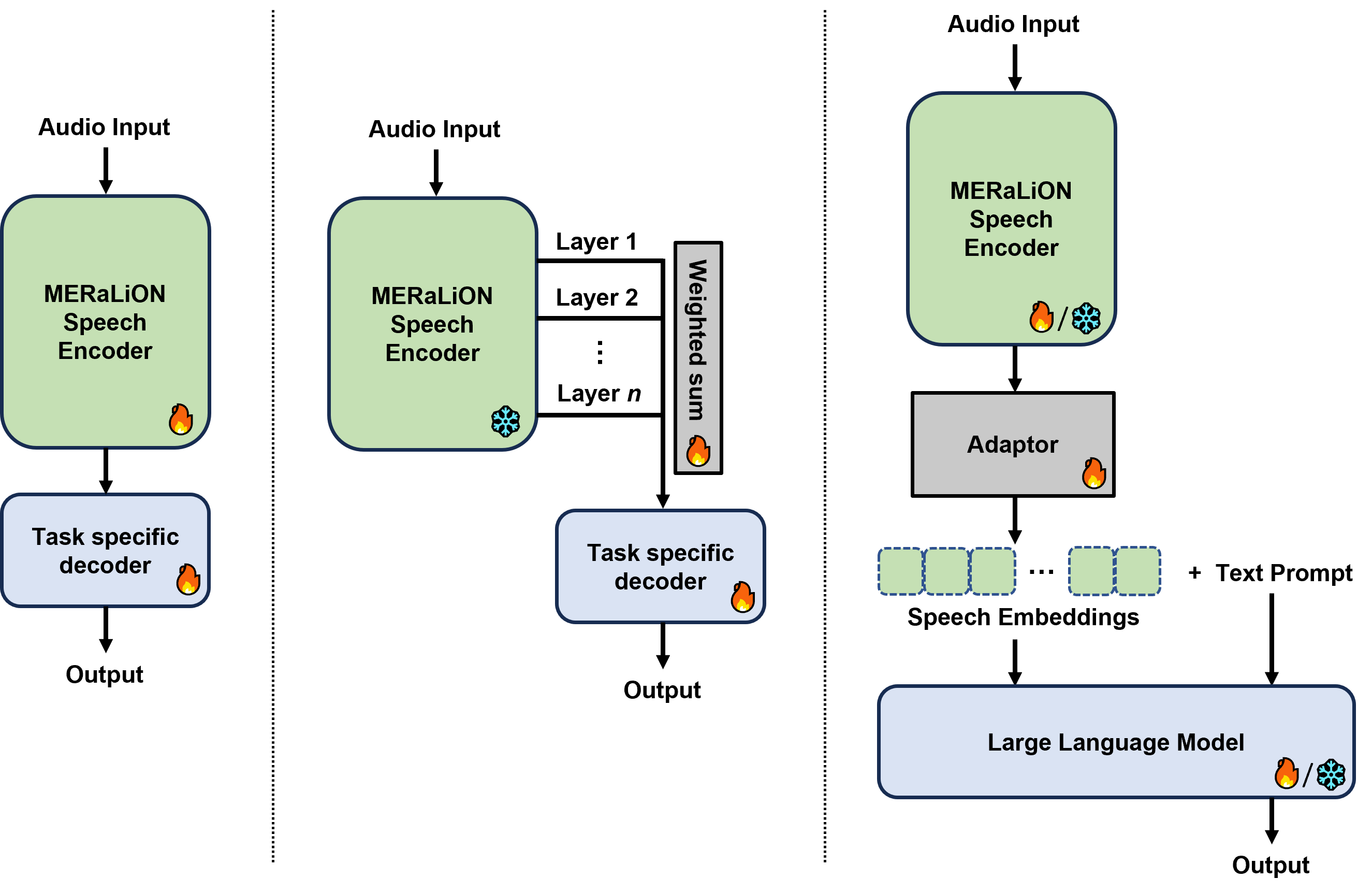}
  \caption{Different ways of utilising the MERaLiON-SpeechEncoder. (Left) Decoder layers are added and the entire model is finetuned for a task-specific objective. For example, for ASR applications, a linear layer could be added and trained with a CTC objective (see section \ref{sec:asr}). (Middle) The speech features are extracted and used as input to a decoder finetuned for a specific task. During this process, the encoder features are fixed and only a weighted sum over them is learned. This is the setup used for SUPERB evaluation (see section \ref{sec:superb}). (Right) Integration of the speech encoder with an LLM, whereby the processed speech features are combined with a text prompt to form a multimodal input. See MERaLiON-AudioLLM \citep{meralion-audiollm} for more details.}
  \label{fig:usecase}
\end{figure}

The two-stage paradigm of first pre-training a representation model on large-scale unsupervised data, followed by finetuning on a downstream task with a relatively smaller amount of labelled data, has revolutionized the approach to many speech problems in recent years \citep{9893562}. This innovation is driven by various self-supervised learning (SSL) techniques \citep{Wav2Vec,Wav2Vec2,hubert,Chen2021WavLMLS,data2vec,bestrq}, which are crucial for modelling useful speech representations during pre-training. SSL leverages the intrinsic characteristics of the input speech to train the model without the need for supervised labels specific to any particular task. This approach aims to create a model capable of computing speech embeddings that are broadly informative across numerous tasks \citep{yang21c_interspeech,tsai-etal-2022-superb} -- a so-called \textit{foundation model}. The MERaLiON-SpeechEncoder can then be utilised in various ways, as illustrated in Figure \ref{fig:usecase}. Note that the examples shown are non-exhaustive.

There have been numerous approaches to SSL for speech, centred around various \textit{pretext} tasks \citep{9893562} or learning objectives, in addition to distinct methods to obtain targets. For instance, Wav2Vec 2.0 \citep{Wav2Vec2} is trained with a contrastive loss to distinguish between positive samples and negative distractors given an anchor representation produced by the encoder. More recent predictive approaches have shown, however, to be more effective and easier to train. The most popular technique leveraged by SOTA models like HuBERT \citep{hubert} and WavLM \citep{Chen2021WavLMLS} rely on a masked language modelling objective popularised by BERT \citep{Devlin2019BERTPO} for text. The model is trained to predict masked sections of the input corresponding to targets derived from \textit{k}-means clustered speech features. For HuBERT, initial speech features, for which the \textit{k}-means targets are computed over, are Mel-frequency cepstral coefficients (MFCCs). These are then substituted for latent features extracted from an intermediate layer of the trained encoder in subsequent steps. Meanwhile, WavLM uses HuBERT latent features to compute the targets from the outset. This type of training encourages the encoding of meaningful latent representations from the unmasked segments while capturing long-range temporal dependencies through the masked predictions, learning both acoustic and linguistic patterns in the process.  

For the pre-training of the MERaLiON-SpeechEncoder, we adopted the BEST-RQ objective, first introduced by \cite{bestrq}. BEST-RQ also uses a masked language criterion but simplifies the target computation pipeline by employing a simple random projection of the input features through a projection layer with frozen weights, which is then compared against random cluster centroids. This avoids the computational cost of performing iterative \textit{k}-means over then entire pre-training dataset. The computational cost of BEST-RQ is further reduced, compared to HuBERT and WavLM, by using the same targets over the entire training process, eliminating the need for multiple iterations of target computation. Despite its more streamlined approach, BEST-RQ and its extension to multiple codebooks in \cite{usmzhang2023google} have shown comparable or better performance against other SSL models, particularly for ASR. As such, variants of the technique have been adopted to build industrial-scale ASR systems, including by Google \citep{usmzhang2023google}, AssemblyAI \citep{Ramirez2024AnatomyOI}, and ByteDance \citep{bai2024seed}. Unfortunately, a BEST-RQ model trained at scale has so far not been released to the public, hindering the ability to reproduce reported results or further develop and enhance the technique. 

We independently implemented the BEST-RQ approach. With the goal of supporting the speech processing ecosystem in Singapore and the surrounding region, we collected and processed a speech dataset comprising around 160K hours of English, 30K hours of multilingual speech, and 10K hours of Singapore-based English that includes code-switching. We report the performance of the model on various ASR benchmarks, targeting different scenarios. For this release, we primarily focus on English performance; while the model may be capable of multilingual processing, this evaluation is left to future work. Additionally, to demonstrate the MERaLiON-SpeechEncoder's generalisability, we extend the downstream analysis to ten SUPERB tasks: ASR, phoneme recognition (PR), keyword spotting (KS), query by example spoken term detection (QbE), intent classification (IC), slot filling (SF), speaker identification (SID), automatic speaker verification (ASV), speaker diarisation (SD), and emotion recognition (ER). We investigate whether training toward targets computed from randomly projected speech representations, which have shown success in ASR, also perform well on these additional tasks.

\section{Self-supervised Pre-training}
In the following sections, we outline the model architecture, data preparation, and training procedure.

\subsection{Masked language modelling and random projection quantisation}

\begin{figure}[t]
  \centering
  \includegraphics[width=0.62\linewidth]{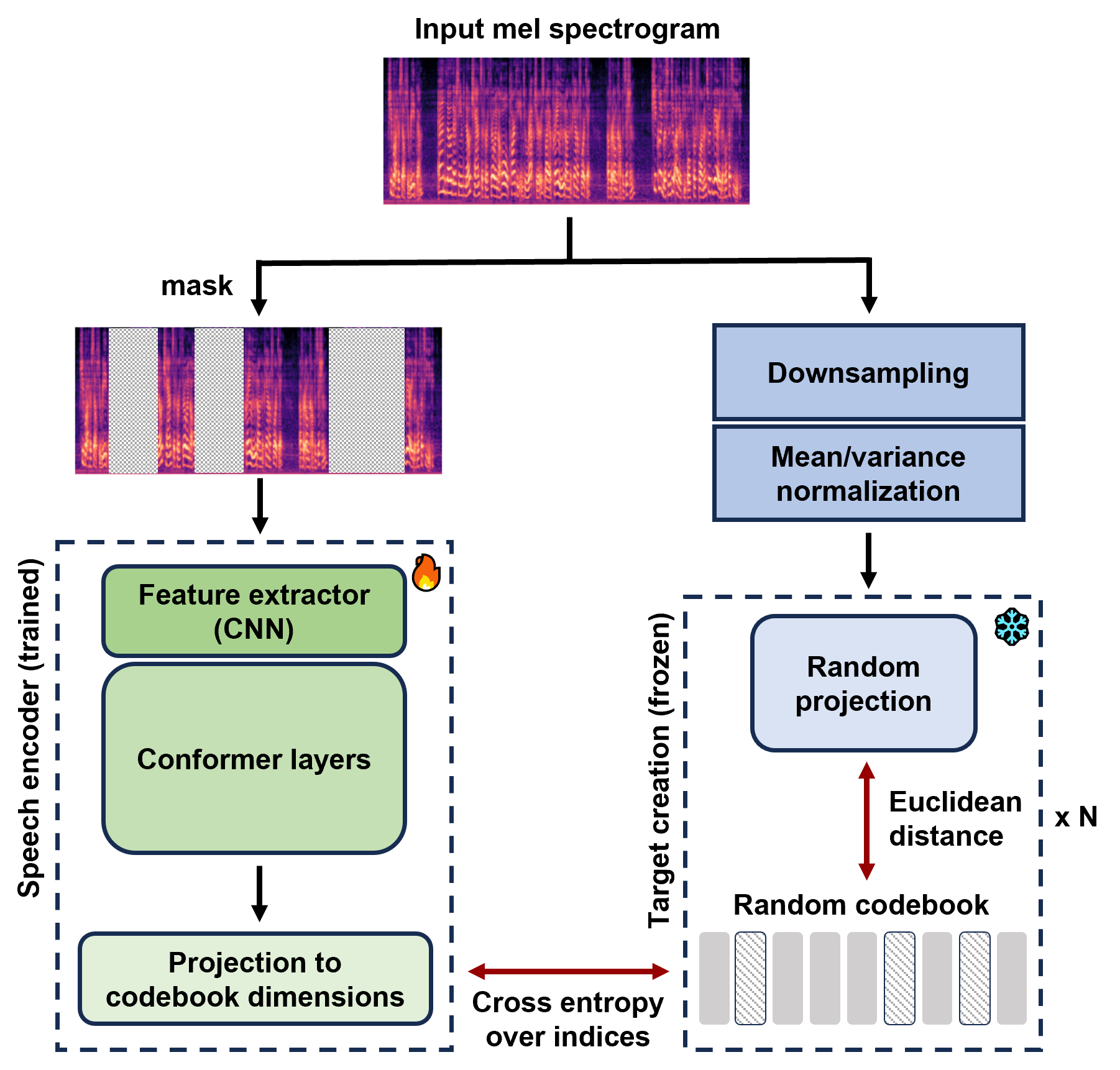}
  \caption{Overview of the pre-training framework. The random projection quantiser comprises a projection layer, to map speech signals to a codebook to create discrete target labels. The speech encoder is trained to predict the labels given by the random projection quantiser over the masked frames of the input speech. After pre-training, only the feature extractor and Conformer layers \citep{gulati2020conformer} are retained as the ``MERaLiON-SpeechEncoder'' which can be further finetuned for downstream applications.}
  \label{fig:best-rq}
\end{figure}

The main pre-training objective follows BERT-style masked-language modelling. This entails predicting the correct discrete label from a codebook, over the masked frames of an input speech signal. Figure \ref{fig:best-rq} provides an overview of this framework. To create the target labels, we apply a projection matrix to the input speech signal, obtaining a frame-wise hidden representation that is subsequently mapped to the nearest vector in a codebook. The index of this vector within the codebook is assigned as the label for that particular frame. Both the projection matrix and codebook are randomly initialised and frozen throughout training. The SSL objective involves training an encoder to predict the labels given by the random projection quantiser from a masked version of the same speech input. The cross entropy loss is computed only over the masked frames. Instead of a projection into a single codebook, using multiple $N$ codebooks together with a multi-softmax loss has been shown to further improve performance for speech translation and reduce variations between different runs \citep{usmzhang2023google}. This may be due to a gradient with smaller variability, arising from the averaging over the multiple codebooks.

The same masking strategy as described in \cite{bestrq} is adopted, whereby each frame of the input speech, represented as a Mel-spectrogram, is masked according to a fixed probability with a fixed span from the starting frame. Overlapping masks are allowed, and the masked sections are replaced with noise sampled from a normal distribution with a mean of zero and a standard deviation of 0.1. The random projection quantiser requires a downsampling mechanism to match the sequence length of the labels with that of the output of the speech encoder. Via PyTorch's \textit{unfold} function, a sliding window was used to group features across frames and stack them in the channel dimension. This feature stacking operation downsamples the target inputs by 4$\times$ to match the downsampling rate of the feature extractor on the encoding side. 

Normalisation of the inputs in the target creation pipeline was found to significantly impact performance. This may be because ensuring an adequate overlap between the distributions of projections and codewords empowers an efficient utilisation of the available codewords. However, unlike the original implementation which uses both a mean/variance normalisation on the inputs and $L^2$ normalisation of the codebook and random projection features, we found that this combination caused extremely slow convergence during pre-training. Using either normalisation independently alleviated this issue, with mean/variance normalisation outperforming $L^2$ normalisation in terms of the pre-training loss and downstream ASR performance. Therefore, our final configuration applied only mean/variance normalisation at the segment level, after the initial downsampling, omitting $L^2$ normalisation. The normalised features, $\bm{x}$, were then projected through a separate random matrix, $\mathbf{A}_j$, for each of the $j$ codebooks. We modified the mapping function between the random projection features $\mathbf{A}_j\bm{x}$ and $i$th codeword $\bm{c}_{ij}$, to an $n$-dimensional Euclidean distance instead of cosine similarity, which proved more stable during training. Here, $n$ is the dimensionality of the codeword vectors. Using this distance, the target label, $y^\text{ref}_j$, was computed as

\begin{align}
  y^\text{ref}_j &= \underset{i}{\text{argmin}} \left(\mathbf{A}_j\bm{x}-\bm{c}_{ij}\right)^\top\left(\mathbf{A}_j\bm{x}-\bm{c}_{ij}\right).
\label{eq:loss}
\end{align}

\subsection{Encoder architecture}

The MERaLiON-SpeechEncoder contains approximately 630M parameters, comprising 24 Conformer layers, with a hidden size of 1024, feedforward size of 4096, convolution kernel size of 5, and 8 attention heads per layer. The Conformer stack is preceded by a two layer convolutional neural network (CNN)-based feature extractor, interspersed between ReLU non-linearities, and a linear layer at the end. Note that the pre-training methodology itself is agnostic to the choice of encoder architecture. 

We utilised an implementation of the Conformer from Fairseq \citep{ott2019fairseq} that includes Multi-head Attention layers with relative positional embedding originating from ESPnet \citep{watanabe18_interspeech}, which was lacking in the Conformer implementation in Torchaudio \citep{yang2022torchaudio}. Preliminary ablation studies showed that the use of relative positional embeddings improved the pre-training loss, compared to other positional embedding methods, including learned, absolute, and RoPE; and all positional embedding strategies outperformed the Torchaudio implementation without positional embedding (see section \ref{sec:relemb}).

\subsection{Pre-training data}
For this release, we utilised a total of approximately 200K hours of unsupervised speech data (i.e. speech audio samples only, without transcriptions or task-specific labels), predominantly in English to pre-train the encoder. To improve the robustness and generalisability of the model, we strove to compile a diverse dataset covering a wide range of conditions, encompassing factors such as domain, style, speaker, gender, and accent. Data was sourced from eight open and publicly accessible datasets, detailed in Table \ref{tab:data_breakdown}. As far as possible, the official training splits of the datasets were used if provided. For VoxPopuli \citep{wang-etal-2021-voxpopuli} and MLS \citep{pratap20_interspeech}, only the English splits were considered. The massively multilingual dataset Common Voice \citep{ardila-etal-2020-common} was included to further increase the diversity of the data although multilingual capability was not a focus this time. To target the commonly spoken variety of English in Singapore, we included the National Speech Corpus (NSC) \citep{koh19_interspeech}. This dataset is notable for not only consisting of Singapore-accented speech, but also containing Singlish terms, often involving heavy use of code-switching, as well as Singaporean named entities. Further details on the NSC dataset, including characteristics of the different parts, are provided in Appendix \ref{sec:appendix_imda}.

\begin{table}[h]
\centering
\caption{Breakdown of data sources used during pre-training and their respective sizes}
\begin{tabular}{lcc}
\toprule
Dataset&Language&Size (hrs)\\
\midrule
Librispeech \citep{panayotov2015librispeech}&English&1K\\
Gigaspeech \citep{chen21o_interspeech}&English&10K\\
VoxPopuli \citep{wang-etal-2021-voxpopuli}&English&24K\\
People's Speech \citep{galvez2021people}&English&30K\\
MLS \citep{pratap20_interspeech}&English&32K\\
Libri-light \citep{kahn2020libri}&English&60K\\
Common Voice 15.0 \citep{ardila-etal-2020-common}& 113 languages&30K\\
National Speech Corpus (NSC) \citep{koh19_interspeech}&English / Code-switch&10K\\
\bottomrule
\end{tabular}
\label{tab:data_breakdown}
\end{table}

We limit the duration of inputs by discarding utterances below 0.3s while randomly cropping those beyond 40s. Cropping was done on-the-fly and was resampled at each epoch. The raw audio data was converted and stored in the Arrow format for efficient loading during pre-training.  

\subsection{Training setup}

We carried out the pre-training in two phases. An initial model was pre-trained on 60K hours of speech from Libri-light. This model and pre-training settings were used to run several preliminary experiments and carry out hyperparameter tuning at a smaller scale prior to the full pre-training, as elaborated in section \ref{sec:prelim}. The final model in this phase was trained for 325K steps on 12 Nvidia A100 40GB GPUs.

Continuous pre-training starting from the above checkpoint was observed to perform better than starting from random initialisation when the dataset was expanded to the full 200K hours. We therefore initialised the encoder with the Libri-light pre-trained model for the full pre-training (see section \ref{sec:cpt}). This training phase was carried out on 128 AMD MI250x GPUs for a further 382K steps, corresponding to about 600 hours, spread over a two-month period. The Adam optimiser was adopted with an inverse square root decay scheduler, using a peak learning rate of 8e-4 and 4K warmup steps. Inputs were batched according to duration using a stratified bucketing schema such that batch size is inversely proportional to duration. This method reduces padding and utilises the GPU RAM more efficiently, leading to improvements in throughput. Further technical details on the use of the AMD GPU cluster and our batching methodology are provided in Appendix \ref{sec:compute}.
%Our method is further detailed in Appendix \ref{sec:batch},  

Following ablation experiments (see section \ref{sec:maskprob}), the masking probability was set to 0.4 for all configurations, differing significantly from the 0.01 used in the original \citep{bestrq}. However, we retained the masking span of 0.4s. Our results also corroborate the benefit of multiple codebooks, discussed in \cite{usmzhang2023google}. We utilised 32 independent codebooks, with a vocabulary size of 2048 each. Each codebook vector had a dimension of 16. Inputs to the model are log-scaled Mel-spectrograms with 80 Mel-bins, a window duration of 25ms, and a shift between windows of 10ms. All audio were resampled to 16kHz prior to the Mel-spectrogram conversion.

\section{Preliminary experiments and ablation studies}
\label{sec:prelim}
In this section we provide details on some of the preliminary experiments carried out to better inform us about the optimal settings for pre-training.

\subsection{Continuous pre-training}
\label{sec:cpt}

\begin{figure}[h]
    \begin{minipage}[b]{0.49\linewidth}
        \centering
        \includegraphics[width=\linewidth]{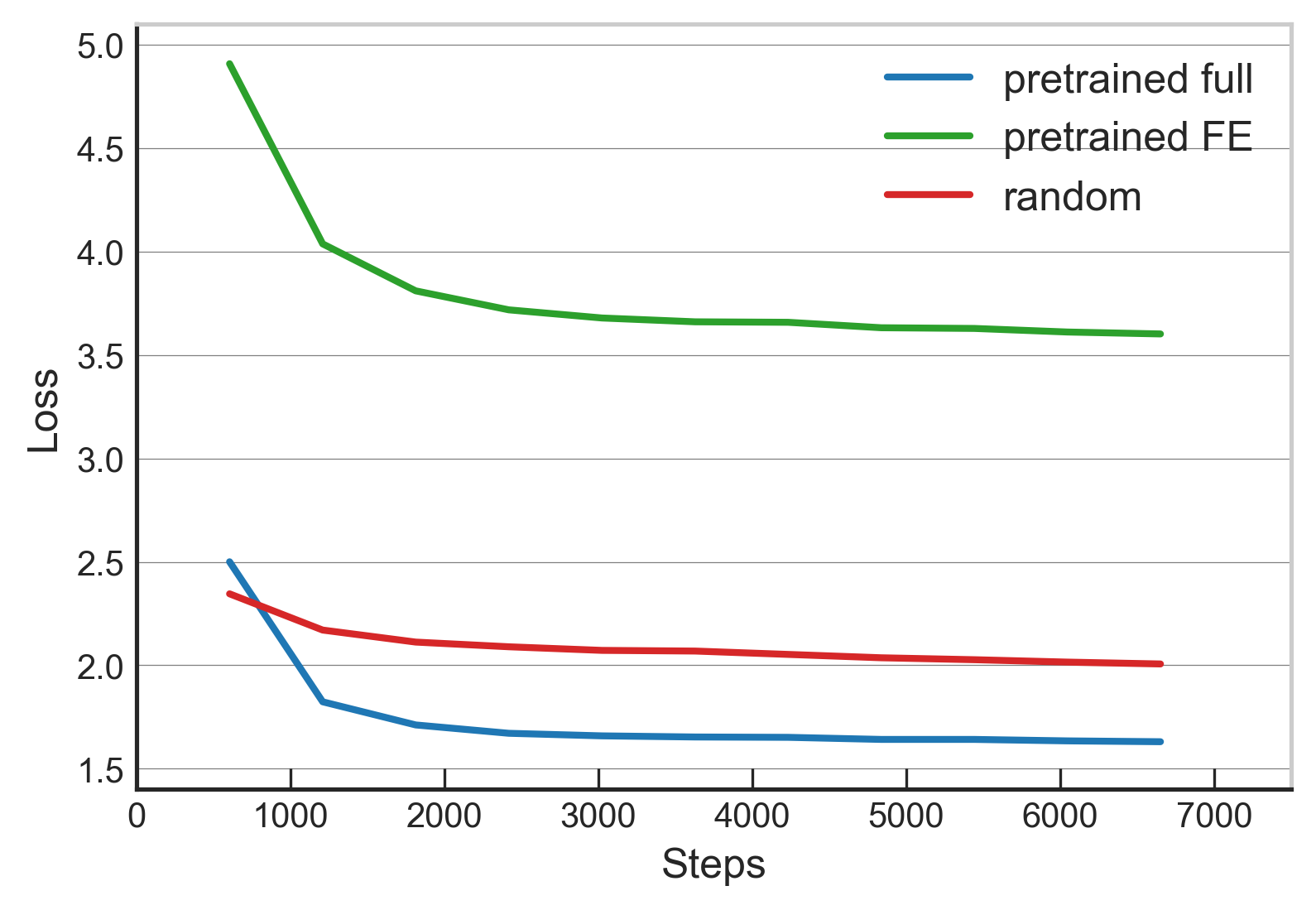}
        \par\vspace{0pt}
    \end{minipage}%
    \quad
    \begin{minipage}[b]{0.49\linewidth}
        \centering
        \includegraphics[width=\linewidth]{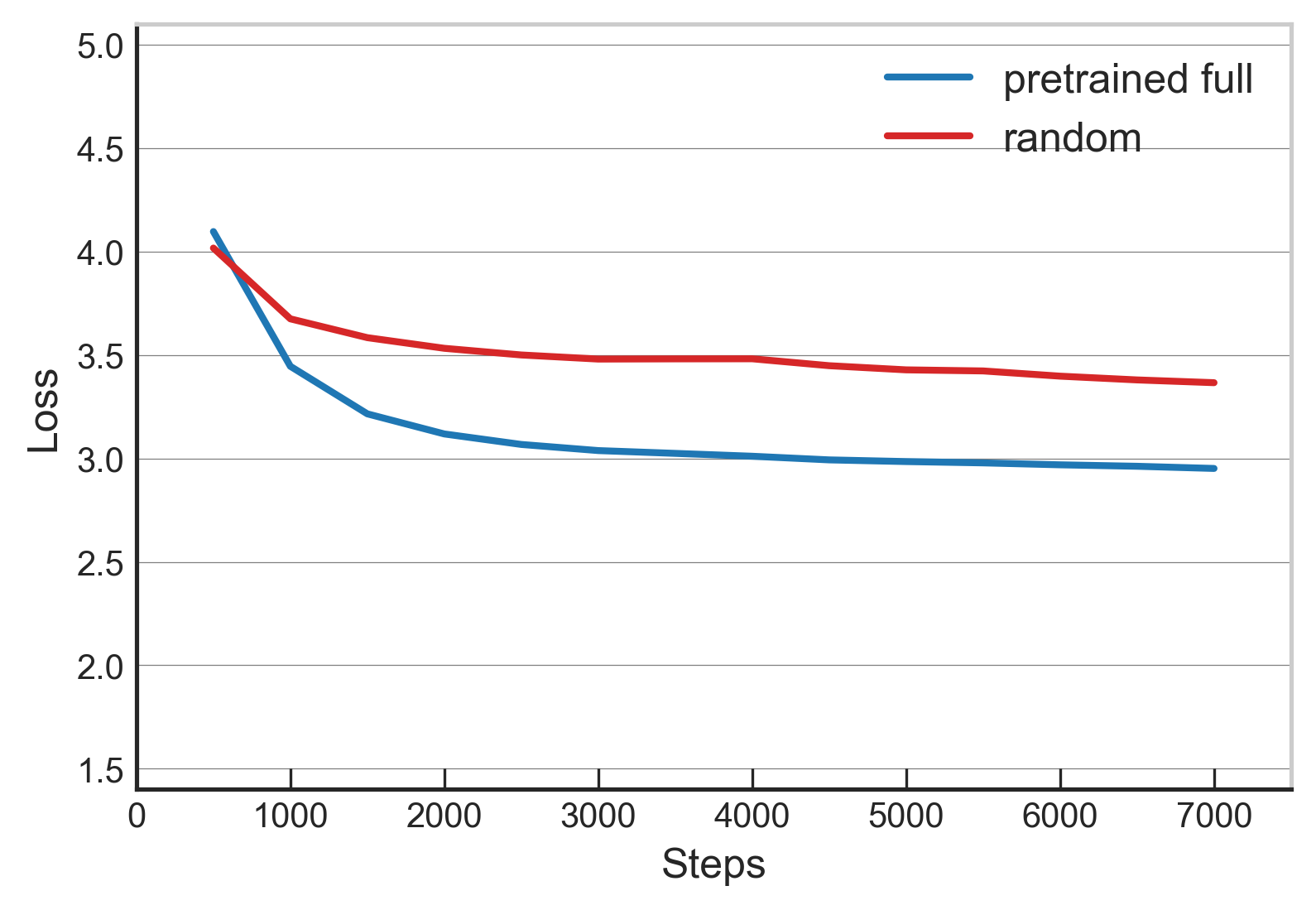}
        \par\vspace{0pt}
    \end{minipage}%
%\end{table}
\caption{(Left) Validation loss on the Librispeech dataset when initialising with a checkpoint previously trained on Libri-light, for all encoder layers (blue) and for only the feature extractor (green), against a randomly initialised baseline (red). (Right) Similarly, the validation curves on the People's Speech dataset when initialised with the same Libri-light checkpoint, for all encoder layers (blue) compared to a randomly initialised baseline (red).}
\label{fig:cont_pre}
\end{figure}

To compare the efficacy between training from random initialisation against continuous pre-training, we compared three different encoder initalisation settings, namely initialising all layers with a primary pre-trained model, initialising only the feature extractor with that from a pre-trained model, and fully random initialisation as a baseline. The encoder pre-trained on Libri-light was chosen as the initial pre-trained model for this experiment. Other parts of the training framework, including the projection quantiser and codebook, were initialised randomly as per usual. To investigate the performance on both in and out-of-domain data, we conducted separate experiments by continuous pre-training on either Librispeech or People's Speech (Figure \ref{fig:cont_pre}); the former of which is, like Libri-light, derived from Librivox audiobooks, while the latter comes from more varied sources.

Continuous pre-training was found to be beneficial, with the models initialised with pre-trained encoders outperforming random initialisation after a few hundred steps, according to the validation loss curves. This was seen even when the domain of the original trained checkpoint did not match the domain of the new training data. Initialising just the feature extractor with trained layers as opposed to the entire encoder, however, significantly degrades performance, and is in fact worse than random initialisation. This suggests a tight coupling between the feature extractor and the Conformer layers, and training them separately may be sub-optimal compared to joint training. Given the above findings, we initialised the full encoder with the Libri-light model for the full pre-training run with 200K hours of data.

\subsection{Positional embedding}
\label{sec:relemb}

We extensively compared various positional embedding methods and implementations. These include relative, learned, RoPE, and absolute, which were all implemented in ESPnet \citep{watanabe18_interspeech}. We also add an alternative Fairseq implementation of absolute positional embedding. The above were contrasted against the Torchaudio version of the Conformer that omits any positional embeddings. Depending on the method, positional embeddings may come in the form of additional parameters or calculations before the Conformer layers, like in the case of learned or absolute, or directly modify the Transformer, like for relative. These changes result in different parameter sizes among the encoders tested, with absolute and RoPE flavours at a consistent size with the baseline at 608M, relative being slightly larger at 630M, and learned being the biggest at 670M. 

\begin{figure}[h]
  \centering
  \includegraphics[width=0.58\linewidth]{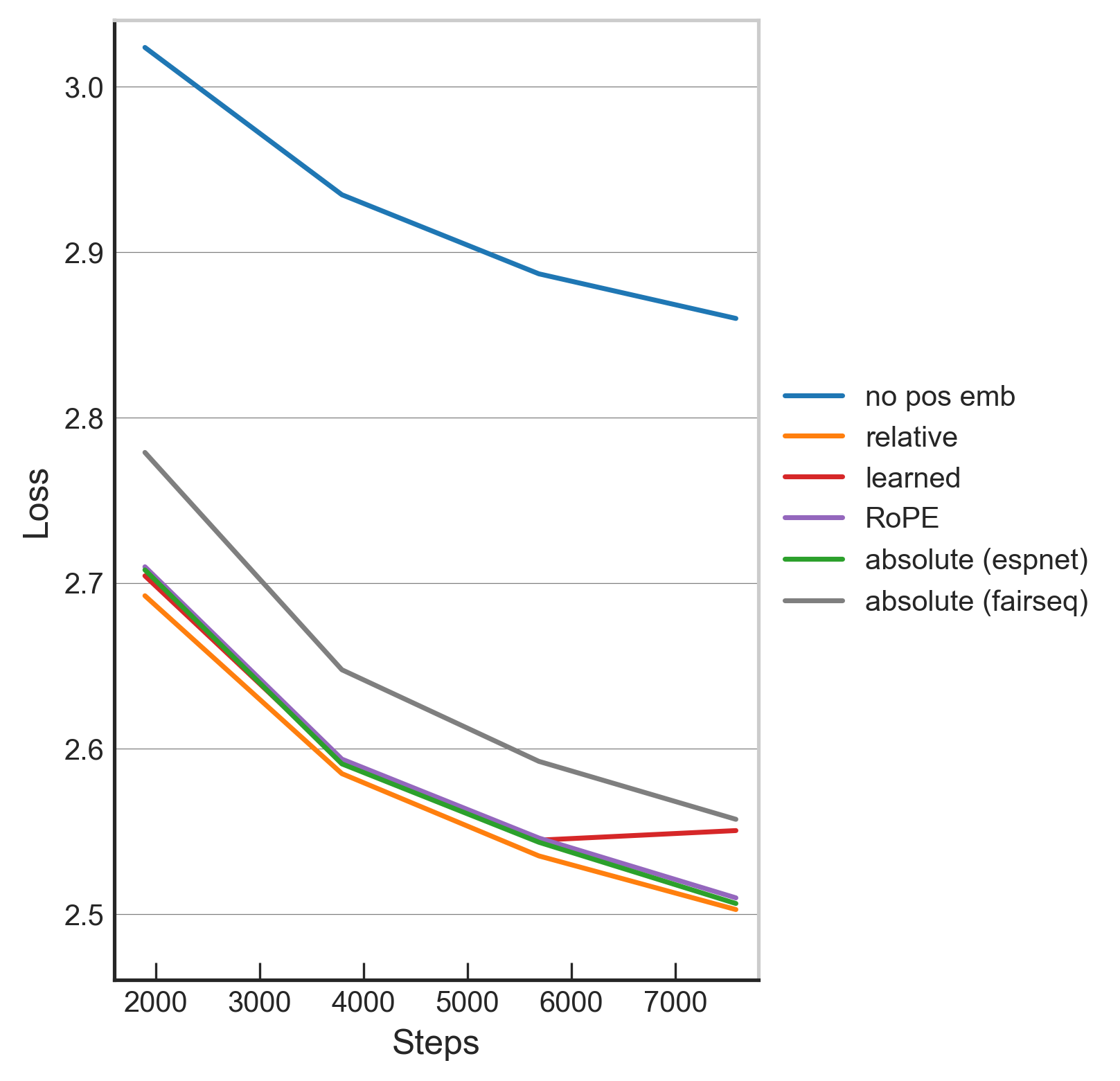}
  \caption{Validation loss over training steps for different types of positional embeddings: relative (orange), learned (red), RoPE (purple), the ESPnet implementation of absolute (green), and the Fairseq implementation of absolute (grey), compared to not using positional embeddings (blue).}
  \label{fig:pos_emb}
\end{figure}

Empirically, we observed slight differences in performance among the various positional embedding techniques. As seen in Figure \ref{fig:pos_emb}, relative positional embedding outperformed the other methods slightly. There is evidently also a gap between the same method implemented differently when comparing the ESPnet and Fairseq versions of absolute positional embedding. Furthermore, it was clear that any positional embedding method was better than not using any. Overall, while we found that models using positional embeddings take longer to train because of extra calculations, we ultimately decided that this penalty was worth the improvements in performance.  

\subsection{Masking probability and codebook configuration}
\label{sec:maskprob}

\begin{figure}[h]
    \begin{minipage}[b]{0.62\linewidth}
        \centering
        \includegraphics[width=\linewidth]{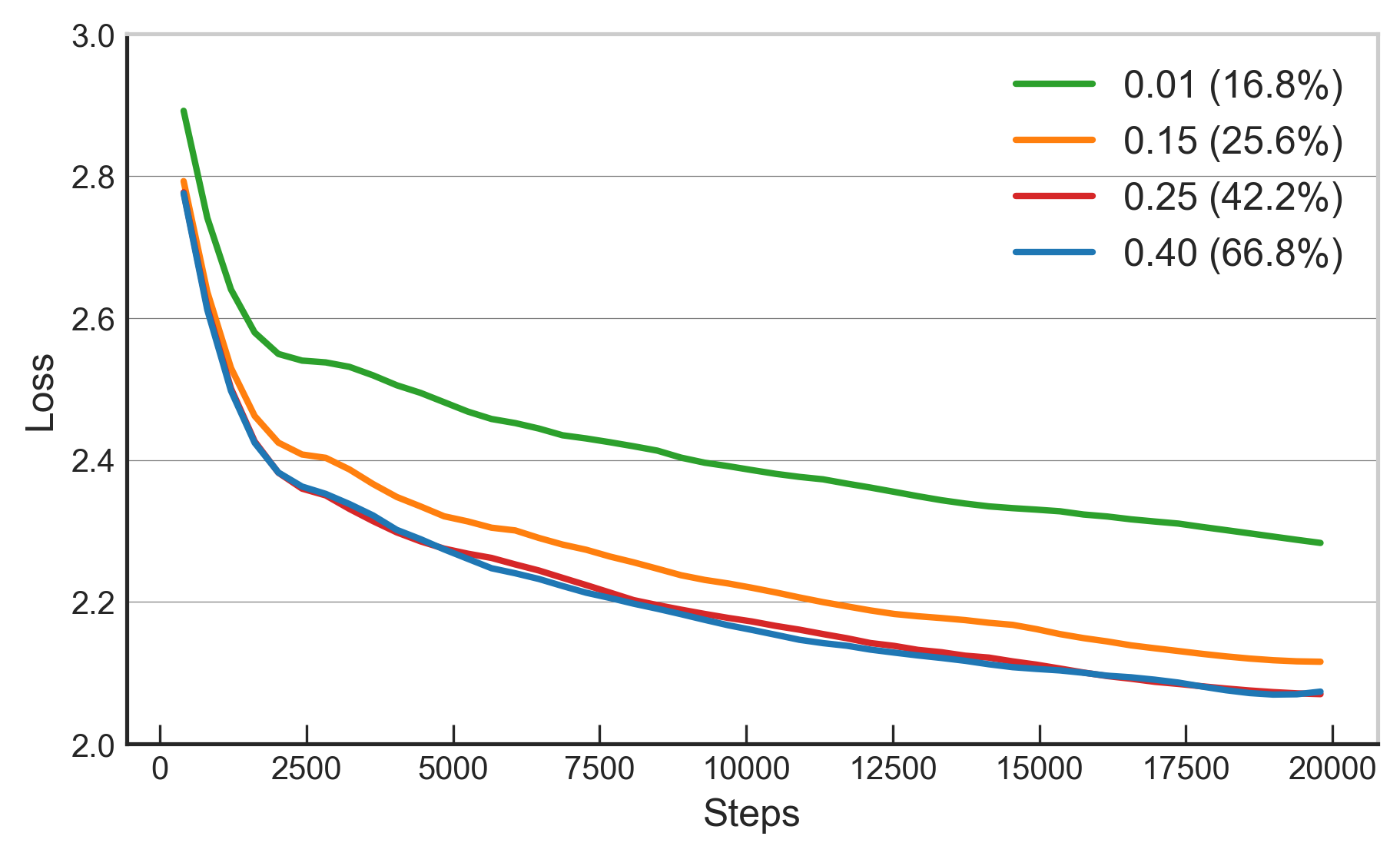}
        \par\vspace{0pt}
    \end{minipage}%
    \quad
    \begin{minipage}[b]{0.30\linewidth}
        \centering
        \tabcolsep=3.0pt
        \begin{tabular}[b]{cc}
        \toprule
        Masking \mbox{prob.} & ASR WER(\%)\\
        \midrule
        0.01&21.64\\
        0.15&17.06\\
        0.25&14.62\\
        0.40&12.73\\
        \bottomrule
        \end{tabular}
        \par\vspace{45pt}
    \end{minipage}
%\end{table}
\caption{(Left) Validation loss over steps for different masking probabilities. Approximate masking coverage of the inputs are given in brackets. (Right) Downstream results on SUPERB ASR task.}
\label{fig:masking}
\end{figure}

Masking probability was varied between 0.01 and 0.4, corresponding to a masking coverage of 16.8\% up to 66.8\% of the input. The pre-training loss improves significantly with higher masking rates, allowing training to converge much faster (Figure \ref{fig:masking}). In terms of training loss, we observe diminishing returns beyond 0.25 as masking coverage approaches the majority of the input, hinting to a sweet spot between having a stronger training signal and maintaining enough unmasked frames for context. Nevertheless, when evaluated on the SUPERB ASR task, performance continued to improve with more aggressive masking beyond 0.25. Hence, we adopted a masking probability of 0.4 for our full pre-training run.

%\begin{table}[h]

%\subsubsection{Impact of codebook configuration}

Work in \cite{bestrq} uses a single set of random codebooks with a vocabulary size of 8192. USM \citep{usmzhang2023google} extends this to 16 random codebooks, presumably with the same vocabulary size of 8192 each, though this is not explicitly mentioned. The vocabulary size is larger than that typically used when the output targets are computed using clustering \citep{hubert}. Perhaps, this choice may have been motivated by assuming that random projections may be less optimal, and may therefore require a larger codebook. Here we experimentally validate the choice of vocabulary size and number of codebooks. The SUPERB ASR and IC tasks were chosen to witness the trends over tasks with labels that operate over different time scales and use downstream models that have different output dimensions. 

%As with the full SUPERB benchmarking, the downstream models were trained using the WavLM Large settings for the learning rate and batch size, suggested in \cite{Chen2021WavLMLS}.
%The BEST-RQ large model is trained from scratch for each hyper-parameter setting. To limit computational cost, each model was pre-trained on Librispeech 960 hrs, and the model checkpoint at the 20000th iteration of training was evaluated. 

\begin{table}[h]
\centering
\caption{Number of codebooks and vocabulary size per codebook, with the corresponding results on SUPERB ASR and IC tasks.}
\begin{tabular}{cccc}
\toprule
\mbox{No.} codebooks&\mbox{Vocab.} size&ASR WER(\%)&IC Acc(\%)\\
\midrule
\multirow{3}{*}{16}&1024&14.83&91.83\\
&2048&14.51&91.09\\
&8192&14.62&92.25\\
\midrule
1&2048&17.89&90.43\\
\bottomrule
\end{tabular}
\label{tab:codebook}
\end{table}

Table \ref{tab:codebook} varies the vocabulary size of each codebook with the number of codebooks fixed at 16, and varies the number of codebooks with the vocabulary size per codebook fixed at 2048. Reducing the vocabulary size to 2048 improves the performance for ASR, but may not be optimal for IC. Using 16 codebooks outperforms using a single codebook consistently for both tasks, validating the observation in \cite{usmzhang2023google}.

\section{Downstream Evaluation}
Our results on ASR and SUPERB benchmarks are reported below.

\subsection{ASR finetuning}
\label{sec:asr}

\subsubsection{Dataset details}
To provide a more well-rounded ASR evaluation, we chose several benchmarking datasets with differing characteristics, summarised in Table \ref{tab:data_summary}. Librispeech \citep{panayotov2015librispeech}, derived from audiobooks, is comprised of read speech with a predominantly American-based accent. TEDLIUM Release 3 (denoted as TEDLIUMv3) \citep{TEDLIUMv3}, extracted from TED talks, may be considered more spontaneous speech. Meanwhile, the NSC dataset \citep{koh19_interspeech} was used for Singapore English and Singlish evaluation. 

\begin{table}[h]
    \centering
        \caption{Summary of datasets used for ASR finetuning.}
    \begin{tabular}{@{}lcl@{}}
          \toprule
       Dataset  &  Duration (hrs) & Speech Type\\
       \midrule
        Librispeech & 100 \& 960  & American read speech \\
        TEDLIUMv3 & 452 & Spontaneous speech \\
        \multirow{2}{*}{NSC subset} & \multirow{2}{*}{420} & Singapore English/Singlish with read, \\
        & & spontaneous and code-switch speech \\
\bottomrule
    \end{tabular}
    \label{tab:data_summary}
\end{table}

The MERaLiON-SpeechEncoder was finetuned for ASR on each of the above datasets separately. For Librispeech, we show results on both the reduced 100 hours and the full 960 hours training sets. The combined set of dev-clean and dev-other was adopted for validation, and the final evaluation was performed on the standard test-clean and test-other sets. For TEDLIUMv3 we used the predefined train, validation, and test splits for the same purposes. We utilised a small subset of the full NSC dataset from our designated training split for finetuning, where a 70 hours portion was randomly selected from each of the six parts, totalling 420 hours. An evaluation subset was created by randomly sampling from each of the six parts. More specifics are provided in Appendix \ref{sec:appendix_imda}. Text normalisation was applied to both TEDLIUMv3 and NSC datasets. We used the text normalisation method from the Kaldi ASR recipe for the TEDLIUMv3 dataset \citep{KaldiTEDLIUM} and Whisper-based text normalisation for the NSC dataset \citep{WhisperTextNorm}.

% The LibriSpeech dataset uses a combined set of dev-clean and dev-other for validation, while evaluation is performed on the standard test sets: test-clean and test-other. The TEDLIUMv3 dataset includes predefined validation and test splits within the dataset. For the NSC dataset, we created a randomly selected evaluation subset by sampling from each of the six parts.

% The MERaLiON-SpeechEncoder was finetuned for ASR using both public English datasets and a Singlish dataset. For the English datasets, we utilized Librispeech \cite{panayotov2015librispeech} and TEDLIUM Release 3 (denoted as TEDLIUMv3) \cite{TEDLIUMv3}. The Singlish dataset was derived from a subset of the NSC dataset that consists of six parts, where a 70-hour portion was randomly selected from each part. Details of the NSC dataset are available in Appendix \ref{sec:appendix_imda}. 
% Text normalization was applied to both the TEDLIUMv3 and NSC datasets. We applied the text normalization method from the Kaldi ASR recipe for the TEDLIUMv3 dataset and Whisper-based text normalization for the NSC dataset. The details of the data sets used in this study are summarized in Table \ref{tab:data_summary}.

\subsubsection{Experimental details}
The models were finetuned using a Connectionist Temporal Classification (CTC) objective \citep{CTC_paper}. A single linear layer was appended after the speech encoder, acting as a classification layer. During finetuning, the encoder was initially frozen to stabilise learning and allow the decoder to adapt, before subsequently training the encoder and decoder jointly. This phased training strategy aligns with established finetuning setups described in \cite{Wav2Vec2}, \cite{hubert}, and \cite{Chen2021WavLMLS}. To ensure effective optimisation, distinct learning rates (LR) and warm-up schedules were applied to the encoder and decoder, following the approach used in \cite{bestrq}. The ASR finetuning hyperparameters are shown in Table \ref{tab:asrfinetune_hyp}. 

\begin{table}[h]
    \caption{ASR finetuning hyperparameters.}
    \centering
    \begin{tabular}{@{}lccccc@{}}
      \toprule
         Dataset	& Encoder LR & Decoder LR &	Warmup steps	&Freezing steps \\
         \midrule
Librispeech 100hrs	& 2e-4 & 2e-3	& 1000	&1500\\
Librispeech 960hrs	&1e-4	& 1e-3 &6000	&12000\\
TEDLIUMv3	& 1e-4	& 1e-3 &3000	&6000\\
NSC subset	& 1e-4	& 1e-3& 3000	&6000\\

    \end{tabular}
    \label{tab:asrfinetune_hyp}
\end{table}

Text tokenisation was achieved using a SentencePiece model \citep{kudo-richardson-2018-sentencepiece}, which is extensively adopted for its subword-based segmentation. A vocabulary size of 1023 was selected for the Librispeech and TEDLIUMv3 datasets, while a larger vocabulary size of 5000 was used for the more diverse NSC dataset to better capture the linguistic variations in Singlish. Data augmentation was performed using SpecAugment, a widely used technique for improving robustness to variability in audio input \citep{park2019specaugment}. Two time masks of width 80 were applied with a masking probability of 0.2, along with two frequency masks of length 27. Decoding utilised a beam search strategy to enhance transcription quality, without leveraging an external language model, ensuring the results reflect the raw model performance. 

As an additional baseline, we finetuned a version of WavLM large on the benchmark datasets in-house. In this case, we used the augmentation method suggested in the WavLM paper \citep{Chen2021WavLMLS} with the exception of layerdrop. We have also included the Whisper model, a versatile ASR system developed by OpenAI \citep{WhisperPaper}. Unlike the other baselines, it has a native encoder-decoder architecture and is trained specifically for speech recognition from the outset. It is designed to handle diverse speech inputs, having been trained on 680K and 5M hours of data for v2 and v3 respectively, enabling it to generalise well across different languages, accents, and domains. In our experiments, we utilised Whisper large v2 and v3 models in both zero-shot and finetuned settings to compare their performance with our MERaLiON-SpeechEncoder.

\subsubsection{Results}
The results on the Librispeech dataset are shown in Table \ref{tab:asr_finetuning}. The MERaLiON-SpeechEncoder demonstrated performance comparable to other leading open-source SSL models on both Librispeech splits. Increasing the finetuning data to 960 hours significantly reduces the word error rate (WER) to 2.1\% (test-clean) and 4.3\% (test-other), that is on par with or better than other SOTA models like Wav2Vec 2.0 and HuBERT when finetuned for the same duration. 

\begin{table}[tbh]
    \caption{ASR finetuning results for Librispeech measured in WER(\%). The ``finetuning'' column indicates the total duration of finetuning data used. With the exception of WavLM, the baseline results were directly retrieved from their respective papers.}
    \label{tab:asr_finetuning}
    %\resizebox{1.0\linewidth}{!}{
%\setlength{\tabcolsep}{2.5pt}
 \centering
    \begin{tabular}{@{}lcccc@{}}
    
    \toprule
         Model & Finetuning (hrs) & test-clean	& test-other \\
         \midrule
       %  BEST-RQ base & LS-960& 100 hrs& 10.55 & 25.54\\
       %  BEST-RQ base & LS-960& 960 hrs& 5.41 & 14.30\\
        % \hdashline[0.6pt/5pt]
        %  Wav2Vec 2.0 base~\cite{Wav2Vec2} & LS-960 & 100 hrs& 6.1  & 13.3 \\
        % Wav2Vec 2.0 base~\cite{Wav2Vec2} & LS-960 & 960 hrs& 3.4  & 8.5 \\
        % \midrule
     %    MERaLiON-SpeechEncoder  & LL-60K & 100 hrs& 4.25 & 9.87\\
     %    MERaLiON-SpeechEncoder  & LL-60K& 960 hrs& 2.73& 6.26\\
         MERaLiON-SpeechEncoder   & 100 & 3.3  & 6.1\\
         MERaLiON-SpeechEncoder  & 960& 2.1& 4.3\\
         % \hdashline[1.0pt/5pt]
         \noalign{\vskip 1mm}  
        \hdashline
        \noalign{\vskip 1mm}  
        % ConformerXL (from scratch) & - & 960 hrs&4.84& 13.07\\
       %  \hdashline[0.6pt/5pt]         
         %Wav2Vec 2.0 large~\cite{Wav2Vec2} & LS-960 & 100h& 4.7  & 9.0 \\
         %Wav2Vec 2.0 large~\cite{Wav2Vec2} & LS-960 & 960h& 2.8  & 6.3 \\
         Wav2Vec 2.0 large~\citep{Wav2Vec2}  & 100 & 3.1  & 6.3 \\
         Wav2Vec 2.0 large~\citep{Wav2Vec2}  & 960 & 2.2  & 4.5 \\
         %HuBERT base \cite{noLMSSLnumbers} & LS-960 & 100h & 6.3 & 13.2 \\
         HuBERT large \citep{noLMSSLnumbers}  & 100 & 2.9& 6.0\\
         HuBERT large \citep{noLMSSLnumbers}  & 960 & 2.1& 4.3\\
        WavLM large (finetuned in-house) & 960  & 2.5 & 4.6\\
        Whisper large v2 (zero-shot) \citep{WhisperPaper} & -& 2.7 & 5.2 \\
         \bottomrule
    \end{tabular}
  % }
\end{table}

Table \ref{tab:ASR_ted} summarises the ASR finetuning results on the TEDLIUMv3 dataset. Our model achieves a relative improvement of approximately 10\% over WavLM on both the validation and test sets. This outperformance highlights our model's ability to generalise better to the TEDLIUMv3 dataset, demonstrating more effective handling of spontaneous speech compared to WavLM. Although zero-shot models like Whisper v2 achieve reasonable performance without finetuning (e.g. 5.2\% WER on LibriSpeech test-other and 4.0\% WER on TEDLIUMv3 test), MERaLiON-SpeechEncoder's finetuning capabilities allow it to achieve lower WERs with supervised training. This highlights the advantage of finetuning when labelled data is available, allowing MERaLiON-SpeechEncoder to surpass zero-shot baselines in accuracy.

\begin{table}[h]
\caption{ASR finetuning results for TEDLIUMv3 measured in WER(\%). Whisper results were retrieved directly from \cite{Ramirez2024AnatomyOI} and \cite{WhisperPaper}.}
    \centering
    \begin{tabular}{@{}lccc@{}}
    \toprule
        Model &  Validation & Test \\
                 \midrule
        MERaLiON-SpeechEncoder & 6.0 & 5.6 \\
        \noalign{\vskip 1mm}  
        \hdashline
        \noalign{\vskip 1mm}  
        WavLM large (finetuned in-house) & 6.7 & 6.2 \\
        Whisper large v3 (zero-shot) \citep{Ramirez2024AnatomyOI} & - & 7.3\\
        Whisper large v2 (zero-shot) \citep{WhisperPaper} & - & 4.0 \\
                 \bottomrule

    \end{tabular}
    \label{tab:ASR_ted}
\end{table}

The experimental results for NSC are provided in Table \ref{tab:ASR_NSC}. Compared to the WavLM model finetuned with the same setup, the MERaLiON-SpeechEncoder consistently achieves lower WER across all six parts of the NSC dataset. Specifically, Parts 2 and 4 of the NSC dataset are particularly challenging, as Part 2 contains numerous named entities, while Part 4 includes spontaneous code-switching. The MERaLiON-SpeechEncoder achieved a 36.5\% and 5.8\% relative improvement against WavLM large for Parts 2 and 4 respectively, demonstrating its effectiveness across different subsets of the dataset. Remarkably, the MERaLiON-SpeechEncoder is able to compete against a Whisper large v3 model finetuned on the full (cleaned) NSC dataset, outright doing better for Parts 4-6 and coming fairly close for Parts 1-3, returning an average WER of 15.2 compared to the finetuned Whisper's 16.9 across all parts. This was despite finetuning on only 5\% of dataset available to Whisper, exacerbated by Parts 1 and 2 making up the majority of the dataset. This clearly illustrates how comprehensive pre-training can help reduce the requirements for downstream finetuning given limited data.
% The Whisper v3 model finetuned on the full NSC dataset achieves lower WERs in some parts, such as Part 1 (4.4\%) and Part 2 (3.8\%). However, these improvements come from fine-tuning on the entire dataset, whereas MERaLiON-SpeechEncoder was fine-tuned on a much smaller subset. This demonstrates that MERaLiON-SpeechEncoder delivers competitive performance even with limited fine-tuning data, highlighting its efficiency and adaptability.

\begin{table}[t]
\caption{ASR finetuning results for the NSC subset measured in WER(\%). The ``finetuning'' column indicates the total duration of finetuning data used. The best results are displayed in bold while the second best results are underlined.}
\resizebox{1.0\linewidth}{!}{
    \centering
    \tabcolsep=3.8pt
    \begin{tabular}{@{}lcccccccc@{}}
    \toprule
        Model & Finetuning (hrs) & Part 1 & Part 2 &Part 3&Part 4&Part 5&Part 6 \\
                 \midrule
        MERaLiON-SpeechEncoder & 420 & 7.2 & \underline{11.5} & 20.4 & \textbf{27.4} & \textbf{14.4} & \textbf{10.4} \\
        \noalign{\vskip 1mm}  
        \hdashline
        \noalign{\vskip 1mm}   
        % \midrule
        WavLM large (finetuned in-house) & 420 & 8.5 & 18.1 & \underline{20.3} & \underline{29.1} & \underline{14.8} & \underline{10.7} \\
        Whisper large v3 (zero-shot) & - & \underline{6.9} & 31.9 & 30.0 & 47.5 & 22.0 & 17.5 \\
        Whisper large v3 (finetuned in-house) & 8169 & \textbf{4.4}&\textbf{3.8} & \textbf{18.8} & 29.8 & 20.6 & 24.0 \\
                 \bottomrule
    \end{tabular}
    \label{tab:ASR_NSC}
    }
\end{table}

\subsection{Universal representation evaluation with SUPERB}
\label{sec:superb}
SUPERB is a collection of downstream speech tasks intended to evaluate the generalisability of the performance of embeddings extracted from a speech encoder. We adhered to the standard SUPERB procedure, where the pre-trained SSL model parameters are frozen. Embeddings from each of the pre-trained model's Conformer or Transformer layers were computed and a separate light-weight downstream model was trained for each task with a learned layer-wise weighted average over these embeddings. Each task's performance was then measured for the cascaded usage of the pre-trained model and respective downstream model. A generalisable SSL model should yield good performance across a variety of tasks. This paper assesses ten tasks across five categories. These are automatic phoneme recognition (PR) and speech recognition (ASR) for recognition, keyword spotting (KS) and query by example (QbE) for detection, intent classification (IC) and slot filling (SF) for semantics, speaker identification (SID), automatic speaker verification (ASV), and speaker diarisation (SD) for speaker, and emotion recognition (ER) for paralinguistics (see Table \ref{tab:superb}). An overall score was also computed for each model, by multiplying QbE results by 100, subtracting error rates from 100 and averaging across all tasks, following \cite{Chen2021WavLMLS}. For SUPERB finetuning, we followed the batch sizes and learning rates contained in Table \ref{tab:superb_hyperparameters}. These hyperparameters were chosen simply by using the best result on the test set, between either the default SUPERB hyperparameters or those chosen for WavLM large in \cite{Chen2021WavLMLS}.

\begin{table*}[htb]
  \caption{SUPERB benchmarking results. The metrics for each tasks are phone error rate for PR, word error rate for ASR, accuracy for KS, IC, SID, and ER, maximum term weighted value for QbE, slot-type F1 and slot-value concept error rate for SF, diarisation error rate for SD, and equal error rate for ASV. ParaL refers to the paralinguistics category. The best results are shown in bold while the second best results are underlined.}
  \label{tab:superb}
  \quad
\centering
\resizebox{1.0\linewidth}{!}{
\begin{tabular}{@{}lc|ccccccccccc@{}}
\toprule
\multirow{3}{*}{Model} &  \multirow{3}{*}{Score} & \multicolumn{2}{c}{Recognition} & \multicolumn{2}{c}{Detection} & \multicolumn{3}{c}{Semantics}     & \multicolumn{3}{c}{Speaker} & ParaL \\ 
\cmidrule(l){3-4}\cmidrule(l){5-6}\cmidrule(l){7-9}\cmidrule(l){10-12}\cmidrule(l){13-13} 
  &  &  \textbf{PR}  & \textbf{ASR}    & \textbf{KS}  & \textbf{QbE}   & \textbf{IC}  & \multicolumn{2}{c}{\textbf{SF}} & \textbf{SID}  & \textbf{ASV}  & \textbf{SD}   & \textbf{ER}              \\
          &     & PER(\%)$\downarrow$  & WER(\%)$\downarrow$  & Acc(\%)$\uparrow$ &  MTWV$\uparrow$  & Acc(\%)$\uparrow$      & F1$\uparrow$    & CER(\%)$\downarrow$   & Acc(\%)$\uparrow$  & EER(\%)$\downarrow$   & DER(\%)$\downarrow$   & Acc(\%)$\uparrow$             \\ 
               \midrule
   Wav2Vec 2.0 large & 80.79 & 4.75  &  3.75 &  96.66 &  0.0489  & 95.28  &  87.11  & 27.31  &  86.14  &  5.65   &  5.62  & 65.64                \\
   HuBERT large & 82.25 &   3.53  &   \underline{3.62}  &   95.29 &  0.0353   & \underline{98.76}  &  \underline{89.81} &  \underline{21.76}  &  90.33  &   5.98 &  5.75  &  67.62               \\
    WavLM large & 84.77 &  \textbf{3.06} & \textbf{3.44}   &  \textbf{97.86} &  \textbf{0.0886}  &  \textbf{99.31}  &  \textbf{92.21} &  \textbf{18.36} & \textbf{95.49} &  \textbf{3.77}  &  \textbf{3.24}  &   \textbf{70.62}              \\ 
   MERaLiON-SpeechEncoder & 82.62 &  \underline{3.14} & 4.16  & \underline{97.63}  & \underline{0.0590}  & 98.60  & 88.99 & 23.89  & \underline{91.09} & \underline{5.18}  & \underline{5.06}  &  \underline{68.02}
   \\
   \bottomrule
\end{tabular}
}
\end{table*}

The results in Table \ref{tab:superb} compare the MERaLiON-SpeechEncoder against SOTA SSL speech foundation models of comparable parameter size: Wav2Vec 2.0 large, HuBERT large, and WavLM large. Wav2Vec 2.0 large and HuBERT large were trained on 60K hours of speech from Libri-light \citep{kahn2020libri}, while WavLM large was trained on 94K hours of speech from the combination of Libri-light, GigaSpeech \citep{chen21o_interspeech}, and VoxPopuli \citep{wang-etal-2021-voxpopuli}. The MERaLiON-SpeechEncoder performs comparably against HuBERT large, and approaches the WavLM performance on several tasks. This suggests that it is still possible to yield comparable performance across a diversity of tasks, even when making the computational saving trade-offs of the random projection targets and the 4$\times$ downsampling. All models outperform Wav2Vec 2.0 large, while WavLM overall performed the best across all tasks.

\section{Future Directions}
%data scaling - targeting non-en
%general improvement of non-asr tasks
%improved representation learning
%integration with AudioLLM
%expanded evaluation
The model released in conjunction with this technical report marks the first version of the MERaLiON-SpeechEncoder. The future roadmap for our foundation model extends the language coverage to major languages spoken in Southeast Asia besides English. In Singapore, we aim to support the other official languages Malay, Chinese, and Tamil. Outside of Singapore, the goal is to gradually include Indonesian, Javanese, Sundanese, Filipino, Thai, Vietnamese, Burmese, Khmer, and Lao. This list is non-conclusive and other languages may also be considered. Accordingly, we are in the process of scaling our data further in preparation for future training runs.

With the expanded language coverage we also wish to increase the breadth and depth of our evaluation to include multilingual benchmarks, like the multilingual(ML)-SUPERB \citep{shi23g_interspeech}, among others. In our view, the original SUPERB itself is showing signs of saturation in terms of results for certain tasks like KS, IC, and ASR, and it may be beneficial to expand the evaluation of these tasks to other datasets. In terms of downstream performance, while better than other SOTA models in several ASR domains, particularly for Singapore English and Singlish, the MERaLiON-SpeechEncoder still falls slightly behind WavLM across the general range of SUPERB tasks. We aim to further refine the training procedure to close this gap, for example through better representation learning during target creation or with the inclusion of augmentation techniques during pre-training. 

In general, future research paths will also be aligned with the MERaLiON-AudioLLM in order to better support the speech modality in a unified system.

\section{Conclusion}
In this report, we present the MERaLiON-SpeechEncoder, a 630M parameter encoder that acts as a speech foundation model to support a wide range of downstream speech tasks. The encoder was trained from scratch in two phases, first on a smaller 60K hours dataset, before expanding to the full 200K hours. We adapt the BEST-RQ SSL technique for pre-training, by leveraging masked language modelling with randomly projected representations as targets. We demonstrate excellent results for ASR on spontaneous speech and Singapore-accented English, as well as Singlish with the inclusion of code-switch. The model is also generally competent across various other speech tasks encompassing the SUPERB benchmark, and holds its own against other SOTA encoders. We hope sharing our model and experiences will be a catalyst for the advancement of speech processing technologies, especially in Singapore and the surrounding region.

%%%%%%%%%%%%%%%%%%%%%%%%%%%%%%%%%%%%%%%%%%%%%%%%%%%%%%%%%%%%
\newpage

\appendix

\section{Appendix}

% \subsection{Efficient variable batching}
% \label{sec:batch}

\subsection{Efficient Training with AMD GPU cluster}
\label{sec:compute}
Our training setup employed a high-performance computing cluster comprising 64 AMD MI250x accelerators across 16 nodes on the LUMI Supercomputer. Each MI250x accelerator is considered as two GPUs from both software and Slurm Workload Manager perspectives due to its dual-GCD (Graphics Compute Die) design. This configuration resulted in a total compute setup of 128 GPUs, with each node hosting 8 GPUs.

Networking between nodes utilised the HPE Cray Slingshot-11 with a 200 Gbps interconnect. Each node was equipped with four endpoints corresponding to the four AMD MI250x GPU modules. Each endpoint facilitated up to 50 GB/s of bidirectional bandwidth, ensuring efficient communication across the network. We leveraged mixed precision training to optimise performance and resource utilisation. Distributed data parallelism (DDP) was employed, treating each GCD as a separate GPU. This allowed our setup to efficiently scale across multiple nodes.

To handle batches with varying sequence lengths and still leverage compilation features, we utilised the automatic dynamic shape compilation introduced in PyTorch 2.1. Our approach involved classifying each sequence into one of six predetermined length buckets and padding only up to the maximum length within each bucket, thus minimising padding tokens while avoiding excessive recompilations due to constantly changing input shapes. Our method achieves a balance between fixed padding to a global maximum (which avoids recompilation given static input shapes but wastes compute on padded inputs) and the standard bucketing approach of padding up to the maximum in a mini-batch (minimal padding but PyTorch recompiles for every unique input shape). Nevertheless, an uneven distribution of sequence lengths introduced loss spikes at epoch transitions, as the model exhibited bias towards more frequently occurring sequence lengths towards the end of an epoch. To address this, rather than iterating through the length buckets at equal intervals, we implemented a sampling-based round-robin dataloader that sampled batches according to their probability of occurrence, proportional to the number of members in each of the buckets.

The training process took approximately 25 days to complete. During instances of heavier node usage, a performance decline was observed due to increased communication initiation time relative to gradient reduction time. To resolve this, we disabled communication bucketing in DDP (note that this is different from the aforementioned length bucket and is part of the DDP configuration) by setting a very large bucket size, effectively minimising overhead and optimising communication efficiency.

\subsection{Details of National Speech Corpus dataset processing}
\label{sec:appendix_imda}
The raw NSC dataset comprises approximately 10,600 hours of recordings of Singaporean English speakers, systematically organised into six distinct parts.
%\begin{itemize}
%\item  Part 1: 3,000 hours of prompted readings based on phonetically balanced scripts.
%\item Part 2: 3,000 hours of prompted readings featuring sentences centered around various topics, including people, food, locations, and brands.
%\item Part 3: 900 hours of conversational data, covering discussions on daily life and gameplay interactions.
%\item Part 4: 900 hours of code-switching conversations where speakers alternate between Singlish and their Mother Tongue languages (Chinese, Malay, Tamil).
%\item Part 5: 1,500 hours of conversations structured around four themes: debate, finance, positive emotion, and negative emotion.
%\item Part 6: 1,300 hours of simulated phone calls across three thematic categories:
%        Holiday, hotel, and restaurant;
%        Bank, telephone, and insurance;
%        Housing and Development Board (HDB), Ministry of Education (MOE), and Ministry of Social and Family Development (MSF).
%\end{itemize}
Although the NSC data set is a valuable resource for model training, it contains a notable amount of mislabelled data, systematic inconsistencies, and accidental errors. To ensure data integrity and reliability, we implemented rigorous verification and filtering procedures, extracting only the most accurate and high-quality segments.

\textbf{Data Splits Consistency}: For Parts 1 and 2, we ensured that examples with identical transcriptions were consistently assigned to the same data splits to prevent data leakage.

\textbf{Timestamp Verification}: For Parts 3 to 6, recordings were selected where the audio duration closely matched the transcription timestamp duration. For conversational audio recorded separately for each speaker, we combined both sides by superimposing their respective audio array representations.

\textbf{Segmentation}: Longer conversations were segmented into shorter units, each with a maximum duration of 30 seconds.

\textbf{Transcription Cleaning}: Non-speech annotations such as <mandarin>, <S>, and (ppb) were removed. However, discourse particles (e.g. [oh]), interjections (e.g. !walao!), and fillers (e.g. (um)) were retained to preserve the natural characteristics of spoken dialogue.

The details of training and testing dataset after filtering is shown in Table \ref{tab:NSC_data_filtered}. Even after filtering, the dataset remains substantially large. To facilitate efficient ASR model comparison, we further selected a 70 hours subset from each of the six parts randomly, resulting in a total duration of 420 hours. This smaller, more manageable dataset enables streamlined training and evaluation of ASR models. We used the same test splits as mentioned in Table \ref{tab:NSC_data_filtered} for evaluation in section \ref{sec:asr}.

\begin{table}[h]
    \centering
    \caption{Duration in hours of each split of the NSC dataset after processing and filtering.}
    \begin{tabular}{@{}lcccccc|c@{}}
    \toprule
        Dataset Split & Part 1 & Part 2 &Part 3&Part 4&Part 5&Part 6 & Total\\
                 \midrule
        Train & 3341.96 & 3150.33& 741.07 & 70.15 & 168.03 & 697.94 & 8169.48\\
        Test & 4.95& 4.04& 7.70& 7.29&6.91&6.76 & 37.65\\
         \bottomrule
    \end{tabular}
    \label{tab:NSC_data_filtered}
\end{table}
\subsection{SUPERB parameters}

\begin{table}[h]
    \caption{Hyperparameters used for different SUPERB tasks.}
    \label{tab:superb_hyperparameters}
    \centering
    \begin{tabular}{@{}lcc@{}}
      \toprule
         Task	& learning rate &	batch size	\\
         \midrule
PR	& 2e-4 &	128\\
ASR	&1e-4	&32\\
KS	& 1e-5	& 512	\\
IC	& 1e-4	& 12\\
SF	& 1e-4	& 128\\
SID	& 1e-4 &	32\\
ASV	&5e-5	&512\\
SD	& 5e-3	&256	\\
ER	& 1e-5	& 32\\
\bottomrule
    \end{tabular}
    \label{tab:asr_hyp}
\end{table}

\section*{Acknowledgements}
We extend our sincere gratitude to Shuo Sun, Huayun Zhang and Xunlong Zou for their purposeful discussions and contributions to data management.

We would like to thank the National Supercomputing Centre (NSCC), Singapore for the resources provided including the use of the Aspire 2A and 2A+ Supercomputing platforms, and the EuroHPC Joint Undertaking and the LUMI Consortium for the use of the LUMI Supercomputer Cluster.

This research is supported by the National Research Foundation, Singapore and Infocomm Media Development Authority, Singapore under its National Large Language Models Funding Initiative. Any opinions, findings and conclusions or recommendations expressed in this material are those of the author(s) and do not reflect the views of National Research Foundation, Singapore and Infocomm Media Development Authority, Singapore.

% \newpage
\bibliography{custom}

\end{document}